\documentclass[journal]{IEEEtran}
\usepackage[left=0.75in, right=0.75in, top=0.75in, bottom=0.75in]{geometry}
\usepackage{cite}
\usepackage{float}
\usepackage{amsmath,amssymb,amsfonts}
\usepackage{hyperref}
\usepackage{graphicx}
\usepackage{textcomp}
\usepackage{xcolor}
\usepackage{graphics} 
\usepackage{epsfig} 
\usepackage{mathptmx} 
\usepackage{times} 
\usepackage{amsmath} 
\usepackage{amssymb}  
\usepackage{algorithm}
\usepackage[noend]{algpseudocode}
\usepackage[mathscr]{euscript}
\usepackage{cite}
\usepackage{caption}
\usepackage{color}
\usepackage{multirow}

\algnewcommand{\IfThenElse}[3]{
  \State \algorithmicif\ #1\ \algorithmicthen\ #2\ \algorithmicelse\ #3}

\def\BibTeX{{\rm B\kern-.05em{\sc i\kern-.025em b}\kern-.08em
    T\kern-.1667em\lower.7ex\hbox{E}\kern-.125emX}}
\begin{document}

\title{Autonomous Navigation via Deep Reinforcement Learning for Resource Constraint Edge Nodes using Transfer Learning}

\author{Aqeel Anwar, Arijit Raychowdhury,~\IEEEmembership{Senior~Member,~IEEE}
\thanks{The authors are with the Department
of Electrical and Computer Engineering, Georgia Institute of Technology, Atlanta,
GA, 30332 USA e-mail: (aqeel.anwar@gatech.edu, arijit.raychowdhury@ece.gatech.edu).}
}

\maketitle
\thispagestyle{empty}
\pagestyle{empty}

\begin{abstract}
Smart and agile drones are fast becoming ubiquitous at the edge of the cloud. The usage of these drones are constrained by their limited power and compute capability. In this paper, we present a Transfer Learning (TL) based approach to reduce on-board  computation required to train a deep neural network for autonomous navigation via Deep Reinforcement Learning for a target algorithmic performance. A library of 3D realistic meta-environments is manually designed using Unreal Gaming Engine and the network is trained end-to-end. These trained meta-weights are then used as initializers to the network in a test environment and fine-tuned for the last few fully connected layers. Variation in drone dynamics and environmental characteristics is carried out to show robustness of the approach. Using NVIDIA GPU profiler it was shown that the energy consumption and training latency is reduced by 3.7x and 1.8x respectively without significant degradation in the performance in terms of average distance traveled before crash i.e. Mean Safe Flight (MSF). The approach is also tested on a real environment using DJI Tello drone and similar results were reported. The code of the approach can be found on GitHub: \href{https://github.com/aqeelanwar/Deep-Reinforcement-Learning-DJI-Tello.git}{\textit{https://github.com/aqeelanwar/Deep-Reinforcement-Learning-DJI-Tello.git}}. \textit{The video of the drone with proposed approach will be uploaded to YouTube.}

\end{abstract}


\section{Introduction}
\IEEEPARstart{O}{ver} the past decade, Unmanned aerial vehicle (UAV) are emerging as a new form of IoT devices being used in varied applications such as reconnaissance, surveying, rescuing and mapping. Irrespective of the application, navigating autonomously is one of the key desirable features of UAVs both indoors and outdoors. Several solutions have been proposed to make drones autonomous in an indoor environment. There has been significant work towards using additional dedicated sensing modalities such as RADAR \cite{kwag2004obstacle} and LIDAR \cite{raimundo2016autonomous}, which provide high accuracy in navigation and obstacle avoidance, thus enabling autonomous flights possible. But when the payload, cost and power is taken into account, such systems are heavy, expensive and power hungry, making them almost impossible to be used in low cost Micro Aerial Vehicles (MAV). Ultrasonic SONAR is a cheap alternative but suffers from lack of accuracy and reduced field of view (FOV). They are also line of sight sensors that need to function in an array to provide a depth map.
On the other hand, over the last decade, there has been significant interest in the use of Neural Network (NN) for various robotic applications. In recent years, reinforcement learning (RL) has been extensively explored for enabling a wide array of robotic tasks. The model-free nature of RL makes it suitable in the problems where little or nothing is known about the environment. RL has been successfully implemented in games and has shown beyond human level performance \cite{silver2016mastering}, \cite{mnih2015human}. However, RL is a data-hungry method and often requires more data compared to other machine learning techniques to generate comparable results. The performance of machine learning algorithms depends heavily upon the complexity of the network and the amount of meaningful data available for training. For a complex task, the deeper the NN, the better the performance. Correspondingly, the  amount of meaningful data scales too \cite{bengio2015deep} until the point where the task is not complex enough given the network architecture and performance starts degrading \cite{hester2017deep}. Training a deeper neural network comes with the cost of increased computation. This makes it challenging to be implemented on a limited resource edge node such as a mobile drone. Simpler NNs with real-time training can be implemented on edge nodes, but this is achieved only by compromising the performance of the underlying application.
So, for an acceptable performance, the network should be deep enough, which comes with:

\begin{figure}[t]
\centering
\includegraphics[width=0.49\textwidth]{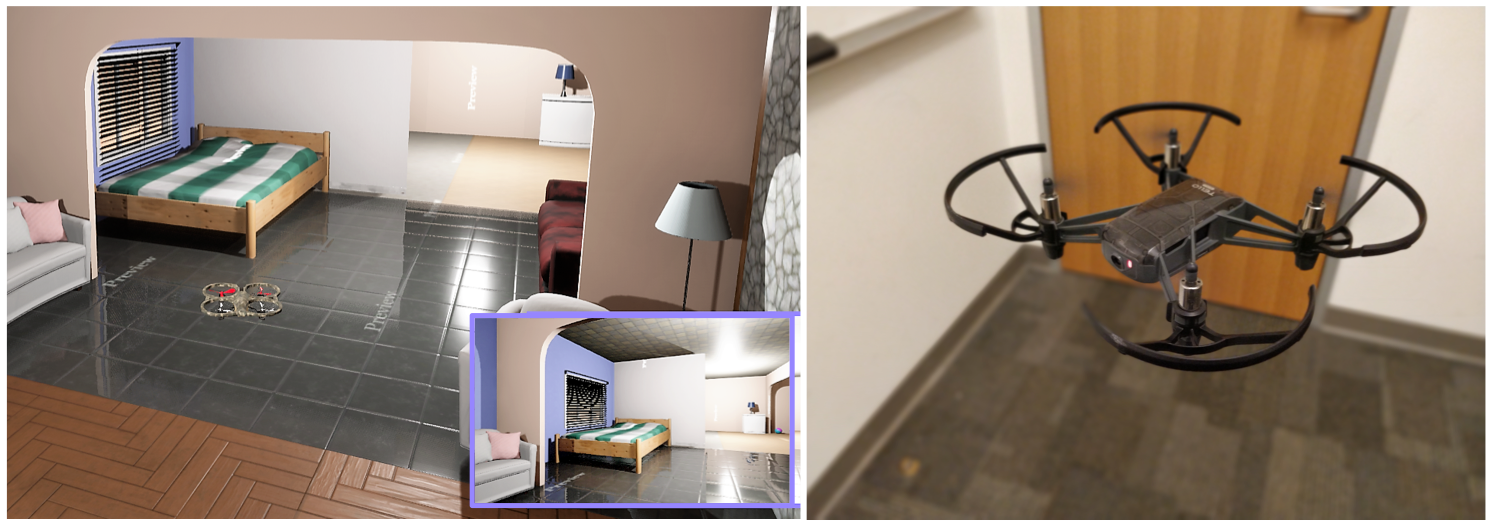}
\caption{(left) DRL for autonomous navigation is carried out an a set of manually generated 3D realistic meta-environments. The learning is transferred to a new test environment and only last few layers are trained. (right) The approach is also tested in a real environment using DJI Tello.}
\label{fig:cover}
\end{figure}

\begin{itemize}
    \item Additional compute requirement
    \item Increased Power consumption
    \item Increased latency
    
\end{itemize}

For a resource constrained edge node (like a light-weight drone), additional compute resource means adding more hardware to the drone decreasing its thrust-to-weight ratio, increased amount of power consumption may drain the battery quicker rendering the drone useless and increased latency will affect its response time making it far from being real-time. Hence these additional requirements are in a direct contrast with drone's inherent limitations. 

Simpler NNs require reduced amount of computations and are possible to be implemented on edge nodes. But for a complex enough task, these simpler NNs do not perform well. So the problem is, for RL related applications how can we implement a neural network training on resource-constrained edge nodes without losing too much performance and with reduced power and latency. One direct approach is to use Offline Training and Deployment i.e. training the NN on cloud, and carrying out inference on the edge nodes. For tasks involving supervised learning (say classification), this is an effective solution. But for Reinforcement Learning (RL) related problems, where there is no clear boundary between the training and inference phase, this can’t be implemented directly. \cite{sadeghi2016cad2rl} however uses an approach where the network is trained on simulated environments posing RL as supervised learning problem and then deployed on new unknown environments. This transfer of knowledge without further fine-tuning doesn't always work well and is tightly tied to the co-relation or similarity between the train and test environments. The more the similarity between the training and testing environment the better the performance and vice-versa. \cite{amer2019deep} learns a CNN with regressors using supervised learning to follow a pre-determined path and fails to perform if the environment changes.  

For the rest of the paper, we will focus on solving autonomous navigation problem using RL in simulated indoor environments.

\section{Related Work}
Since the overall objective is to make Micro Aerial vehicles (MAV) capable enough of carrying out ML training algorithms, this problem can be approached in either of the two areas. The first and more direct approach is to make better hardware engines for DNN accelerators \cite{sze2017efficient, hill2017deftnn}. Authors of \cite{rebecq2017real} design and implement an energy-efficient accelerator for visual-inertial odometry (VIO) that enables autonomous navigation of miniaturized robots. \cite{palossi201964mw} demonstrates a navigation engine for autonomous nano-drones which is capable of closed-loop end-to-end DNN-based visual navigation. The other approach is to devise better and improved algorithms that take lesser amount of computations (hence energy) for similar performance such as model compression \cite{jiaoptimizing, denton2014exploiting}. 
\cite{han2015learning} developed Network Pruning, which begins with a pre-trained model, then the network parameters which are below a certain threshold are replaced with zeros forming a sparse matrix, and finally performs a few iterations of training on the sparse CNN. The downside of this approach is that the network needs to be iteratively pruned and re-trainined until the desired compression is achieved. Moreover this approach might not be useful for online ML problems such as RL where re-training the network is not energy efficient at all. \cite{iandola2016squeezenet} presents SqueezeNet, a CNN architecture that has 50x fewer parameters than AlexNet and maintains AlexNet-level accuracy on ImageNet by exploring the design space of convolutional network. This tiny network might be problem specific and is not guaranteed to be complex enough for convoluted task such as end-to-end autonomous navigation.
This paper proposes an approach that falls in the latter category.

Transfer learning is a well established approach of transferring any prior domain knowledge to a new problem or domain. This is how human brain works, instead of learning any new problem from scratch, it uses pre-existing knowledge about prior problems and uses that along with learning new skill set to solve the problem. Transfer learning has been widely used in Machine Learning problems to address the issues of smaller or insufficient amount of data, mitigating convergence issues, reducing the time/steps required for convergence \cite{taylor2009transfer, da2016transfer, pan2009survey, weiss2016survey,tan2018survey, george2017deep, taylor2007cross, du2019improving}. These issues are addressed by learning a neural network for one task, and using the learned weights as initialization to another network for a different task. The network weights are then fine-tuned based on the new domain knowledge (data-set). The most common and simplest example of TL is using Imagenet learned weights as initializer for classification problems.

To the best of our knowledge all the TL papers in the past discuss TL as tool/approach to address the above-mentioned issues without worrying much about the computational cost required to train a deep neural network. In this paper we show we can use Transfer learning, to segment a deep network into trainable and non-trainable part reducing the training computations, for underlying task without compromising too much on its performance.

\begin{figure}[t]
\centering
\includegraphics[width=0.4\textwidth]{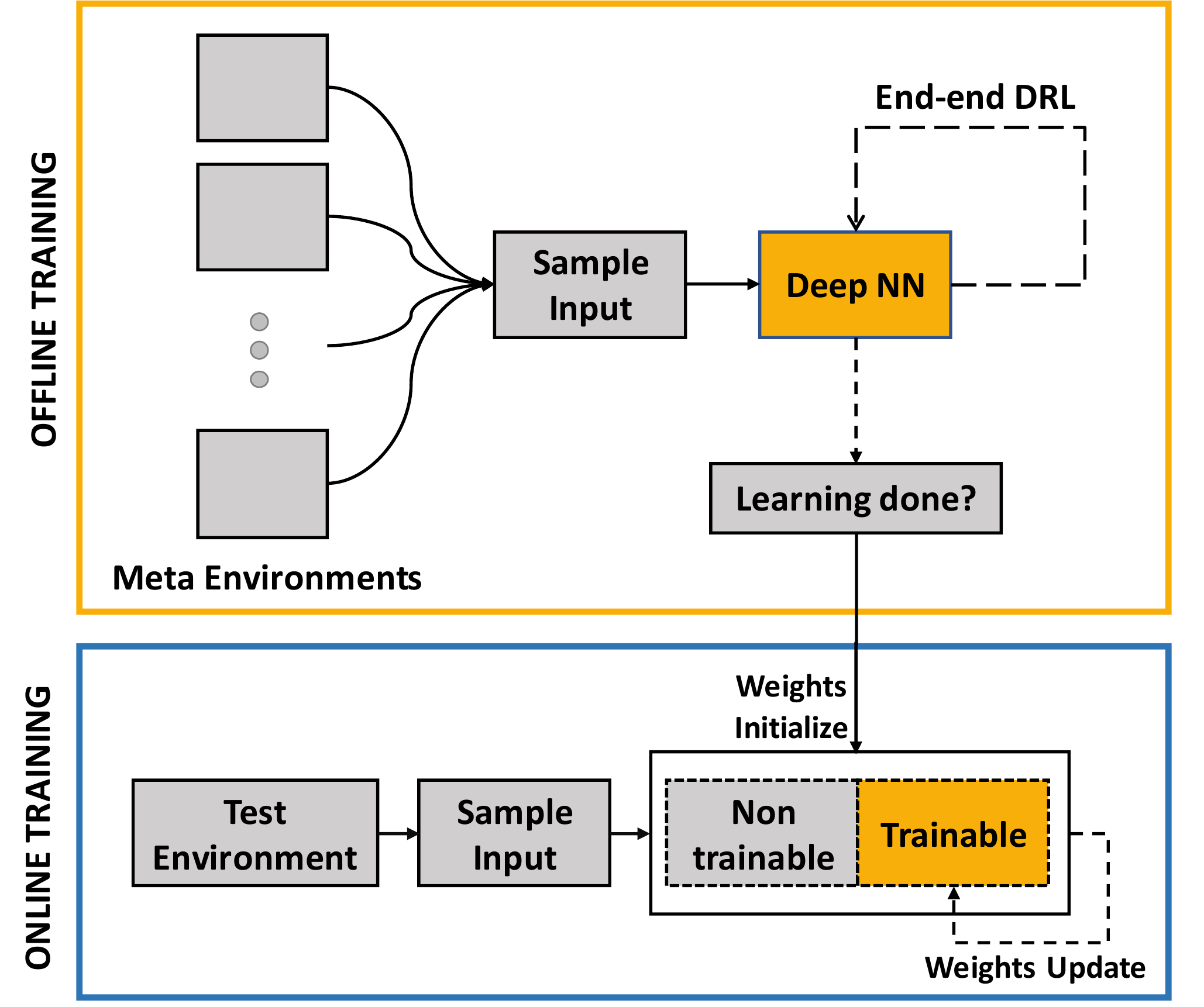}
\caption{Block Diagram for the TL based Approach to DRL}
\label{fig:overview}
\end{figure}

\section{Background on Reinforcement Learning (RL)}

In Reinforcement Learning (RL), the agent interacts with the given environment learning a control policy to achieve the underline objective. As opposed to Supervised Learning (SL) where the target labels are static, the RL training labels are dynamic until the mapping converges. The dynamic nature of the labels (or Q values) requires constant interaction with the environment and can't be done offline. In this paper, the RL objective is to achieve autonomous flight, taking actions that lead to a collision free flight of the drone. There is no predefined start or end position and the goal is to keep on moving across the environment.

Consider the above mentioned task of obstacle avoidance. The agent interacts with the environment $\mathscr{E}$ in a sequence of actions, observations and reward calculations. At each time instant $t$, the agent observes the current camera frame $s_t$. It takes an action $a_t$ from a predefined action space $\mathscr{A}$ and implements it. Implementing the action moves the drone to a new position where it observes a new camera frame $s_{t+1}$. This new camera frame along with the action taken will quantify a reward $r_t$. This reward should be high if the drone moved in the right direction avoiding the obstacle and low if the action took it closer to the obstacle, increasing the chance of collision. Hence each iteration in RL generates a data-tuple $(s_t, a_t,r_t, s_{t+1})$. 
%
The goal for RL is to learn a control policy $a = \pi(s)$ that predicts actions given the state in a such a way that the long term reward is maximized. At each time step $t$, action $a_t$ needs to be predicted that eventually leads the agent to a sequence of states $s_i$ with rewards $r_i$ for $i\in \{t+1, t+2, ... \}$ such that the future discounted return $R_t = \sum_{i=t}^{T} \gamma^{i-t}r_i$ is maximized, where $\gamma \in [0,1]$ is the discount factor. At a given time step $t$, the drone only gets to observe the current frame $s_t$ and hence the task of obstacle avoidance is partially observed. The system can be safely assumed to be a Markov Decision Process  (MDP) where the current state only depends on the previous state and the action taken.

Each of the state-action pair is assigned a Q-value $Q(s,a)$. This Q-value quantifies the expected discounted return achieved by taking an action $a$ at a state $s$ i.e. $Q(s, a) = \mathbb{E}_{\pi} [ R_t | s_t=s, a_t=a ]$.  The idea is to learn these mapping from all the possible states to all the available actions in the action space. This expression when simplified, yields the following Bellman optimality equation
\begin{equation}\label{eq:bellman}
    Q(s,a) = r +\gamma max_{a'} Q(s', a')
\end{equation}

Bellman equation is used to update the Q-values during training. The training data consists of states as input and their corresponding Q values as target output. Once the mapping is effectively learned,  it ensures that in a given state $s_t$ selecting an action $a_t = max_{a'} Q(s_t, a')$ i.e predicting the action with the largest Q value will result in maximizing the future discounted reward $R_t$. 

In Deep Reinforcement learning (DRL), this mapping from states to Q-values $s \longrightarrow Q(s,a)$ is done by learning a Neural Network and hence requires a lot of training iterations before it can converge. Learning to avoid obstacles from monocular RGB images is a complex task and requires deeper neural networks. Training these deep neural networks usually adds to the latency and energy requirements.

      

\section{TL based Proposed Approach}
\subsection{Objective}
Transfer the learning from Cloud to edge nodes for Deep Reinforcement Learning (RL) applications. In this paper we discuss transfer learning based algorithmic improvement targeting
\begin{itemize}
    \item Real-time Learning: Improved training latency
    \item Energy efficient: Reduced energy consumption
    \item Similar performance: No significant degradation in algorithmic performance
\end{itemize}

\begin{figure}[t]

\centering
\includegraphics[width=0.4\textwidth]{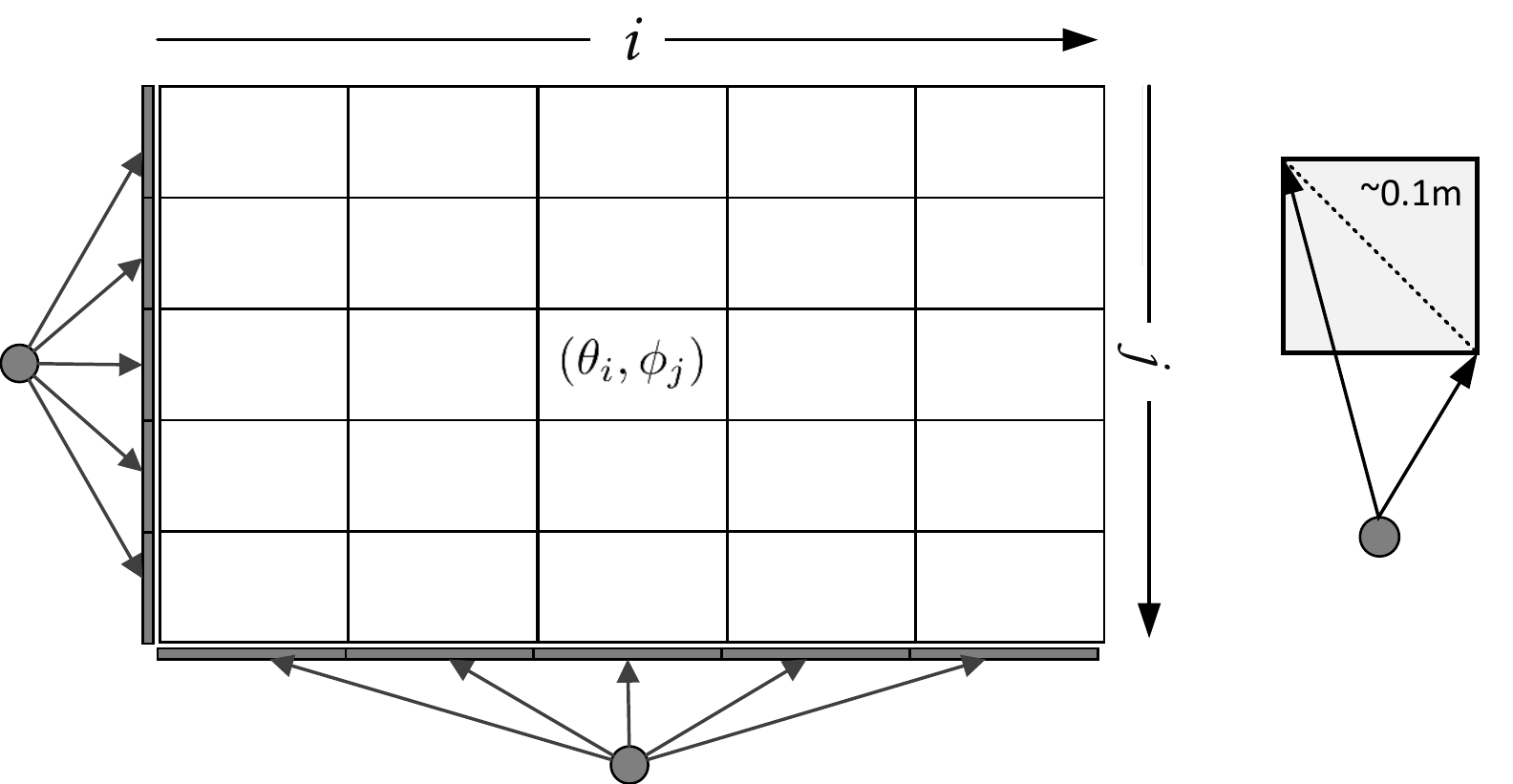}
\caption{Perception based probabilistic Action Space $A_s$}
\label{fig:action_space}
\end{figure}

\subsection{Overview:}
We propose a two-phase approach to the problems related to DRL which combines offline and online learning using Transfer Learning and fine-tuning. The idea is that if we train a NN for an RL application (say autonomous navigation) in a variety of indoor environments collectively, we can use this knowledge using Transfer Learning while training a smaller part NN for similar application in a similar (but different/unseen) test environment. The top-level block diagram of the approach can be seen below. In the Offline phase, one single network is trained on a set of training environments (called meta-environments) using DRL. These environments serve as a library of environment for the underlying problem. This offline training phase is carried out on server (and not on edge-nodes) where we assume no strict restriction on the compute engine. Once we have effectively trained a network on the meta environments collectively, we use these meta-weights as initialization during the online training phase. In the online training phase, a different test environment is used for training (fine-tuning). The training computations need to be carried out in the edge nodes (we don't implement anything on hardware, rather we provide the compute statistics and compare them with training the network end-to-end). In this phase, the training is only carried out on a part of the network. The network is divided into non-trainable and trainable part and only the weights of the trainable part are updated. The segmentation of the network is a compromise between the performance (obstacle avoidance) and the number of training computations. Training the convolution (CONV) layer takes up much more computation as compared to that of fully connected (FC) layer. Also, CONV layers capture the top level features of the underlying problem such as edge detection, blurring and sharpening and as we go deeper into the network, the features become more and more specific to the underlying problem. Hence including the CONV layers within the the non-trainable part of the network makes much more sense. The trainable part of the network consist of last few FC layers. The number of layers in the trainable part of the network is a parameter (called train type) that we vary during the experimentation. The variation of these train types is done by keeping the following two parameters in mind:
\begin{itemize}
\item \textbf{Similar performance:} For the reduced trainable size of the network, we ideally want it to perform similar to that of training the entire network (end-to-end training or e2e). The higher the similarity between the meta-environments and the test environment, the better the performance while training a smaller number of NN layers.
\item \textbf{Reduced Training Computations:} With the reduced trainable weights, we want the training computations to be significantly lower to that of e2e train type. This reduced computations will make the approach practical to be used on resource constraint edge-nodes.
\end{itemize}

\begin{figure}[t]

\centering
\includegraphics[width=0.48\textwidth]{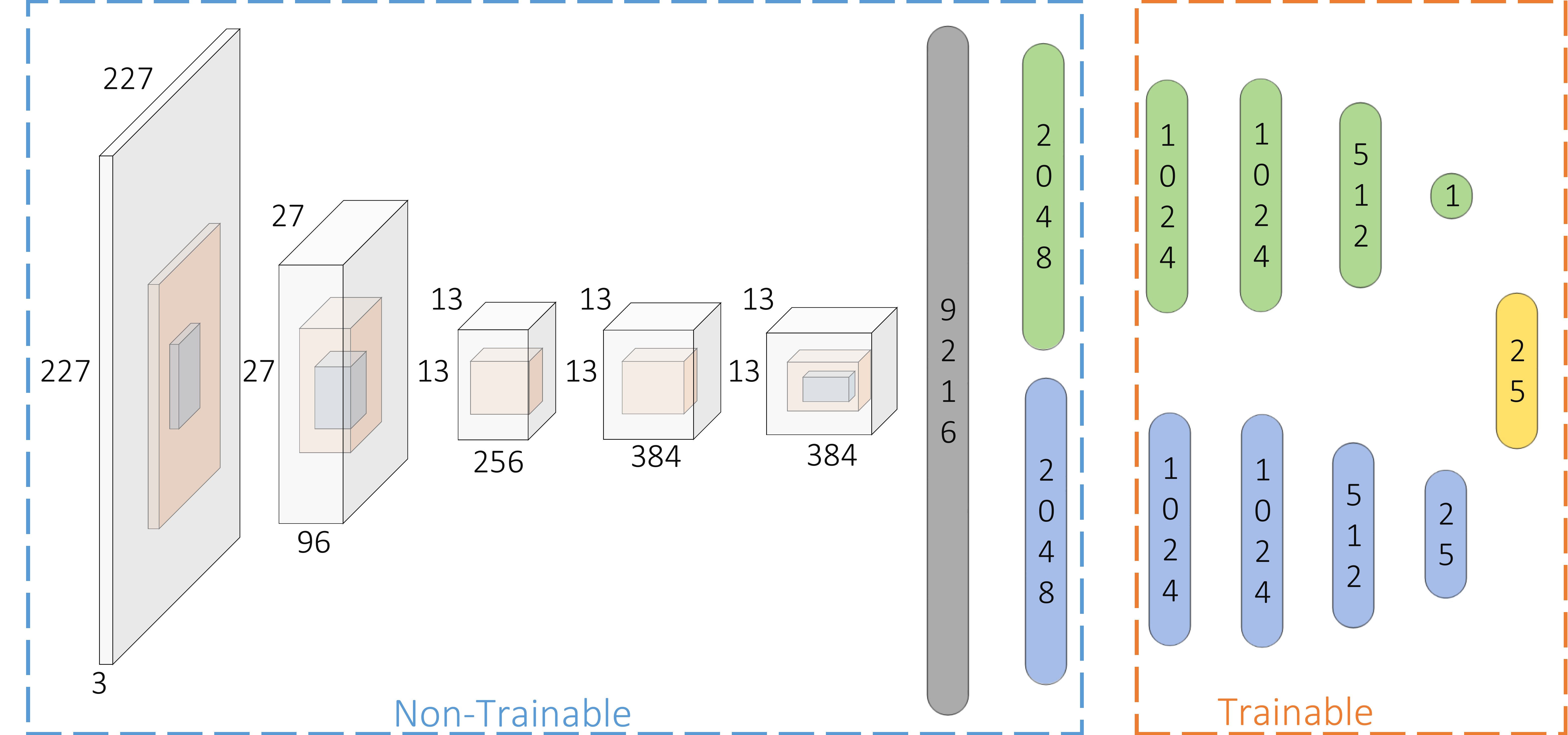}
\caption{Modified Alexnet used for training}
\label{fig:cnn}
\end{figure}

\begin{table}[]
\centering
\caption{Weights and FLOP for each train type}
\begin{tabular}{|l|c|c|c|c|}
\hline
\textbf{Train Type} & \multicolumn{1}{l|}{\textbf{\# of weights}} & \multicolumn{1}{l|}{\textbf{FLOP}} & \multicolumn{1}{l|}{\textbf{\% weights}} & \multicolumn{1}{l|}{\textbf{\% FLOP}} \\ \hline
e2e                 & 48,858,522                                  & 5.16G                               & 100\%                                       & 100\%                                     \\
last4               & 7,358,490                                   & 7.35M                               & 15.06\%                                     & 0.14\%                                    \\
last3               & 3,162,138                                   & 3.15M                               & 6.47\%                                      & 0.06\%                                    \\
last2               & 1,062,938                                   & 1.06M                               & 2.17\%                                      & 0.02\%                                    \\ \hline
\end{tabular}
\label{tab:train_type}
\end{table}

\begin{figure*}[t]

\centering
\includegraphics[width=0.98\textwidth]{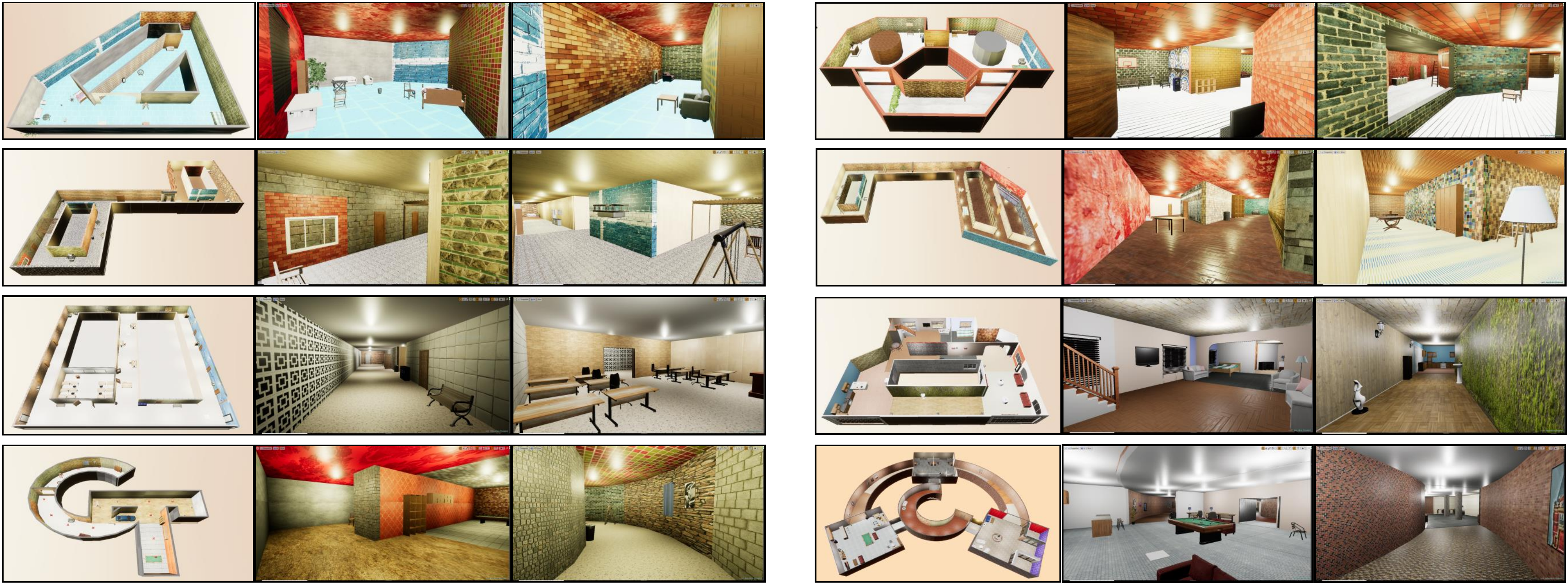}
\caption{3D floor plan and screen-shots of the 8 meta environments used for offline training phase.}
\label{fig:meta_envs}
\end{figure*}

\begin{figure*}[t]

\centering
\includegraphics[width=0.98\textwidth]{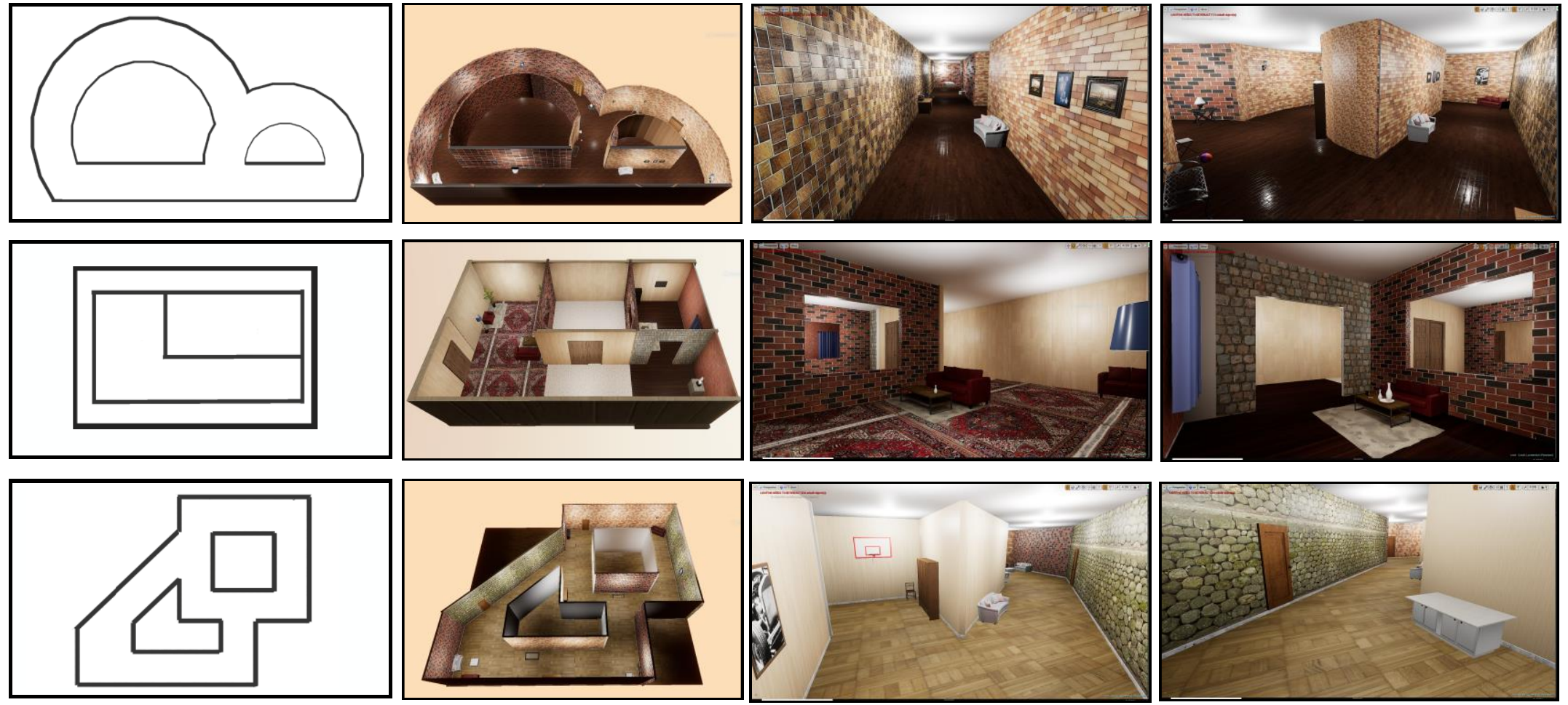}
\caption{3D floor plan and screen-shots of the 3 test environments used for online training phase. (from left to right) Cloud, Condo and Twisty}
\label{fig:test_envs}
\end{figure*}




\subsection{Perception based probabilistic action space $A_s$}
Perception based discrete action space $A_s$ of size $N\times N$ is used. In this action space the agent navigates by controlling the yaw and pitch expanding over all three coordinates. These angles are calculated by making use of the horizontal and vertical field-of-view (FOVs) of the front facing camera. The camera image at time $t$, $s_t$ is divided into $N\times N$ grid. Each window in the grid corresponds to an action in the action space. The action selection is simply the choice of the bin which is then transformed into velocity commands $v_t$ for the drone. This velocity command results in moving along the line connecting the current position to the position where the window becomes the entire camera frame by $r$ meters.
Varying the pitch $\phi$ only results the agent moving to one of the vertical bins while varying the yaw $\theta$ only moves the agent along the horizontal bins. Given the vertical and horizontal FOV ($FOV_{v}$ and $FOV_h$) these angle are calculated as follows 

\begin{equation*}
    \theta_i~ = \left( \frac{FOV_h}{N^2} \times (i-(N^2-1))/2) \right)
\end{equation*}

\begin{equation*}
    \phi_j~ = \left( \frac{FOV_v}{N^2} \times (j-(N^2-1)/2)\right)
\end{equation*}
where $i, j \in \{0, 1, \dots, (N^2-1)/2\}$ is the $(i,j)$ bin location as shown in Fig \ref{fig:action_space}.

In all these actions, the agent moves forward by a constant distant of $r=0.5m$. Moreover the control associated with the action space is probabilistic. A uniform random noise $\epsilon \sim uniform(-b, b)$ is added to these deterministic yaw and pitch angles making them probabilistic and robust to slight control variations where $b=\frac{1}{15}$ is empirically selected. The maximum difference in final position under this probabilistic space for the same action is $\sim 0.1m$ and can be seen in Fig \ref{fig:action_space}

\subsection{Network Architecture}
Deep Neural Network is used to map the state to their corresponding Q values based on a modified Alexnet architecture \cite{krizhevsky2012imagenet}. This architecture takes as input an RGB frame of size $227\times 227\times 3$ and outputs $N^2$ number of  Q-values corresponding to each action in the action space. The network architecture can be seen in Fig \ref{fig:cnn}. In order to help deep reinforcement learning converge better a dueling nature of the network \cite{wang2015dueling} was used where we train two streams of FC network to estimate the state value function $V(s_t)$ and advantage function $A(s_t, a_t)$ separately which can be seen in the figure. Training approach used DoubleDQN \cite{van2016deep} and Prioritized Experience Replay (PER) \cite{schaul2015prioritized} to avoid the over-fitting nature of Bellman Equation and aid faster learning respectively.

The complete network is trained during the offline phase while for the online phase a part of the network is used for training. Extra FC layers are added to the network to quantify the effect of training certain number of layers in the online phase. A Parameter $train~type$ is defined based on the number of layers that are trained. We evaluate the training for 3 different train types denoted by last$p$ and compared to the baseline of training the network end-to-end (e2e) where $p \in \{2,3,4\}$ denotes the number of FC layers trained from the end. 

The idea behind these $train~types$ is that training fewer number of layers will result in reduced computational cost. The details for these train type (number of weights, amount of Floating Point Operations FLOP) can be seen in Table \ref{tab:train_type}. For modified Alexnet architecture, training the last$p$ layers for $p \in \{2,3,4\}$ results in significant reduction in the number of floating point operations required. This reduction in computations is directly co-related to the amount of energy required for training and is reported quantitatively in  in the Section \ref{sec:compute_cost}.

\begin{figure}[t]

\centering
\includegraphics[width=0.48\textwidth]{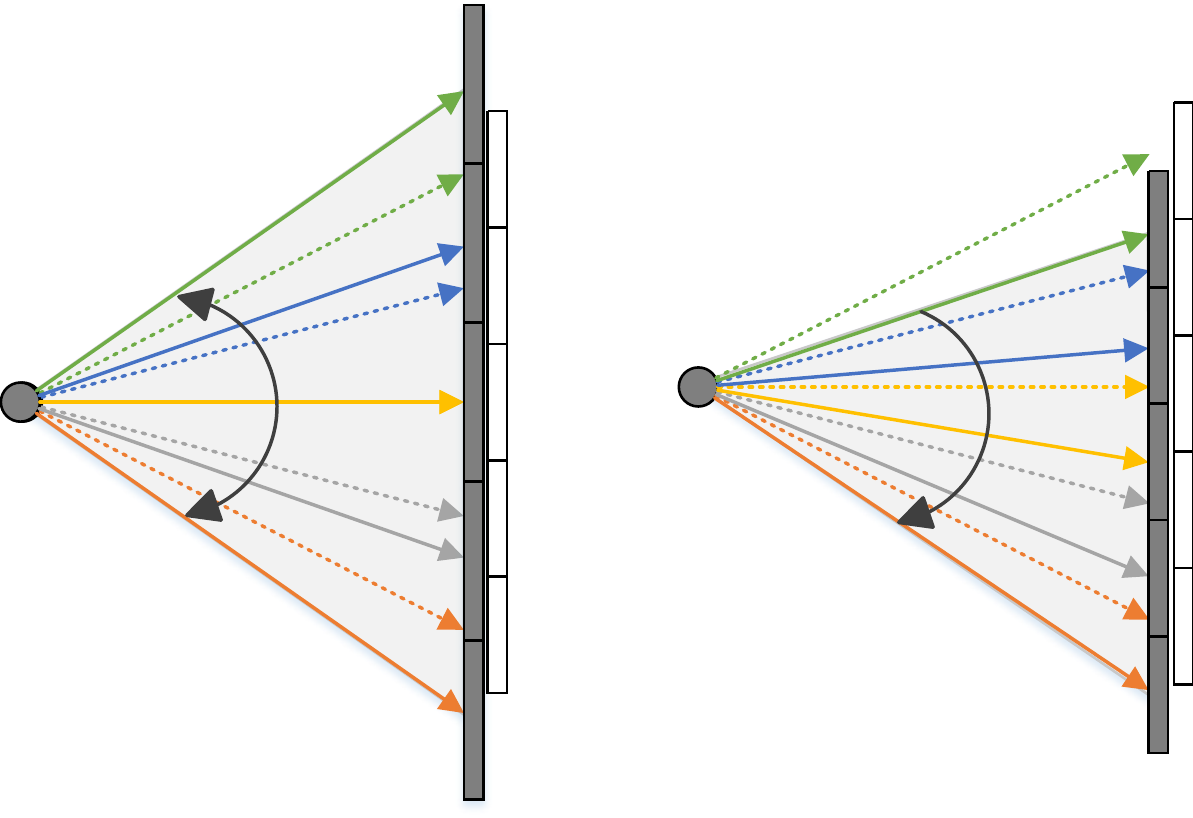}
\caption{Variation in Action Space $A_s$. Right: Rotated Left: Dilated. }
\label{fig:action_var}
\end{figure}

\begin{algorithm}[t]
\caption{\textsc{Offline Training Phase Algorithm}}
\begin{algorithmic}
\State \textbf{Input:} Set of N meta environments: $\mathscr{E}_{meta} = \{\mathscr{E}_0, \mathscr{E}_1, ..., \mathscr{E}_N,\}$
\State \textbf{Output:} Weights of neural network $\theta_{meta}$ 
\State \textbf{Initialization:} Behaviour network: $Q_{\theta}(s) = \mathscr{N}(s;\theta)$, Target network: $Q_{\theta'}(s) = \mathscr{N}(s;\theta')$, $n_{target}$: Target network update interval, $n_{batch}$: mini-batch size for training, $n_{train}$: Train Interval, $\mathscr{D}_{replay}$, $env=0$, $m:$ Environment switch interval

\For{$t \in \{1,2,3,..., max\_steps\}$}
    \If{$mod(t,m)=0$}
        \State $saved\_state[env] \gets (s_t, p_t)$
        \State $env \gets mod((env+1),N)$
        \State $(s_t, p_t)\gets saved\_state[env]$
        \State $\mathscr{E}_{current} \gets \mathscr{E}_{env}$
        \State $position\_agent(\mathscr{E}_{current}, p_t)$
    \Else 
        \State $s_t \gets get\_state(\mathscr{E}_{current}, p_t)$
    \EndIf

    \State Sample an action $a_t$ from current policy using $\epsilon$-greedy
    \State $p_{t+1} \gets move\_agent(\mathscr{E}_{current}, p_t, a_t) $
    \State $s_{t+1} \gets get\_state(\mathscr{E}_{current}, p_{t+1})$
    \State $r_t \gets get\_reward(s_t, a_t, s_{t+1}, p_{t+1})$
    
	\State Store the tuple $(s_t, a_t, s_{t+1}, r_t)$ in $\mathscr{D}_{replay}$
    \If{$mod(t, n_{train})=0$} 
        \State Sample a mini-batch of size $n_{batch}$ from $\mathscr{D}_{replay}$
    	\State Train the Behaviour network: $Q_{\theta}(s) = \mathscr{N}(s;\theta)$
    \EndIf
	\If{$mod(t, n_{target})=0$} $\theta' \gets \theta$
    \EndIf

\EndFor
\State $\theta_{meta} \gets \theta$
\end{algorithmic}
\label{alg:E2ERL}
\end{algorithm}

\subsection{Simulated 3D environments:}
We manually designed all the 3D indoor environments used for experimentation. These environment were built using an open source gaming engine called Unreal Engine \cite{unreal}. The designed environments contain a large variety of lighting conditions, hallway sizes and structures such as long, broad, narrow, sharp turns and circular hallways. Indoor furniture objects with various sizes were used to furnish these environments. The walls were textured with various patterns including metal, wood, marble, concrete and wallpapers. These patterns were selected randomly from a pool of 40 textures to create a diverse data set. Learning a network on this wide variety of indoor environments will help us generalize it to other rendered environments. The more the variation of parameters in the simulation, the better the network is able to generalize the problem. The floor plan and screenshots of the 8 meta-environments can be seen on Fig. \ref{fig:meta_envs}.

The deep learning framework used is TensorFlow. The environments are interfaced with Python using a AirSim plugin \cite{shah2018airsim}. It is an open-source simulator developed by Microsoft for agents (drones and cars) with physically and visually realistic simulations.

\begin{figure}[t]

\centering
\includegraphics[width=0.4\textwidth]{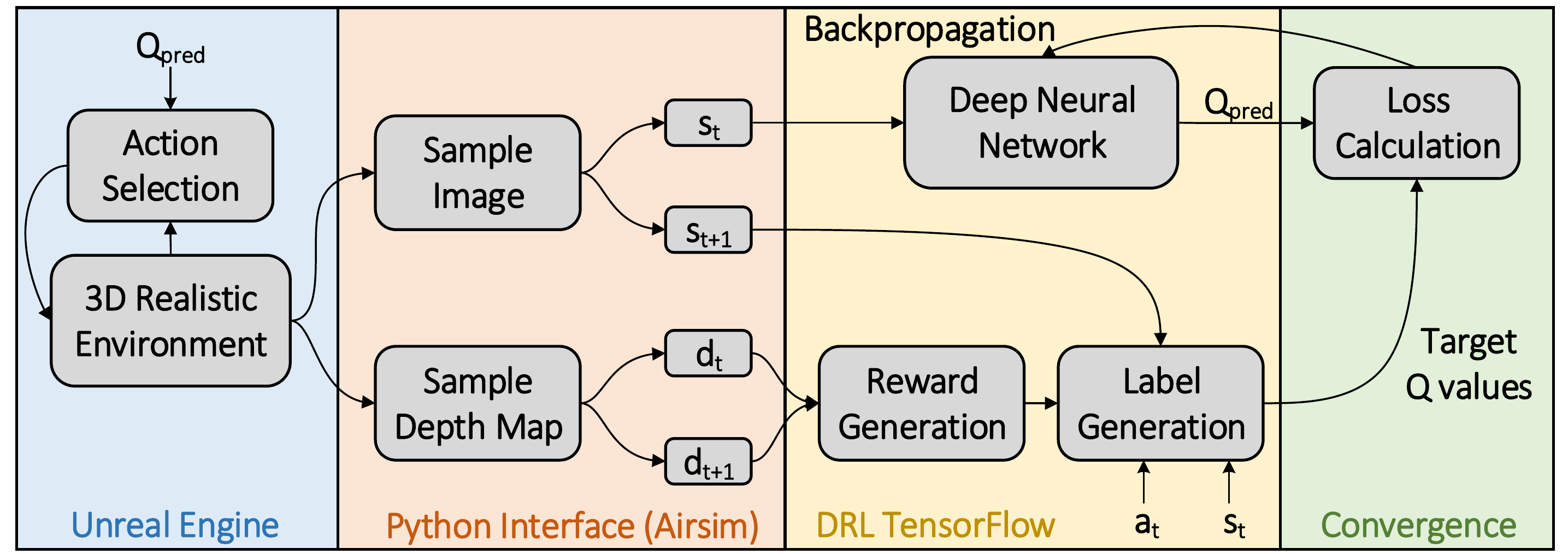}
\caption{DRL Training Block Diagram}
\label{fig:drl_train}
\end{figure}

\begin{figure*}[t]
\centering
\includegraphics[width=0.98\textwidth]{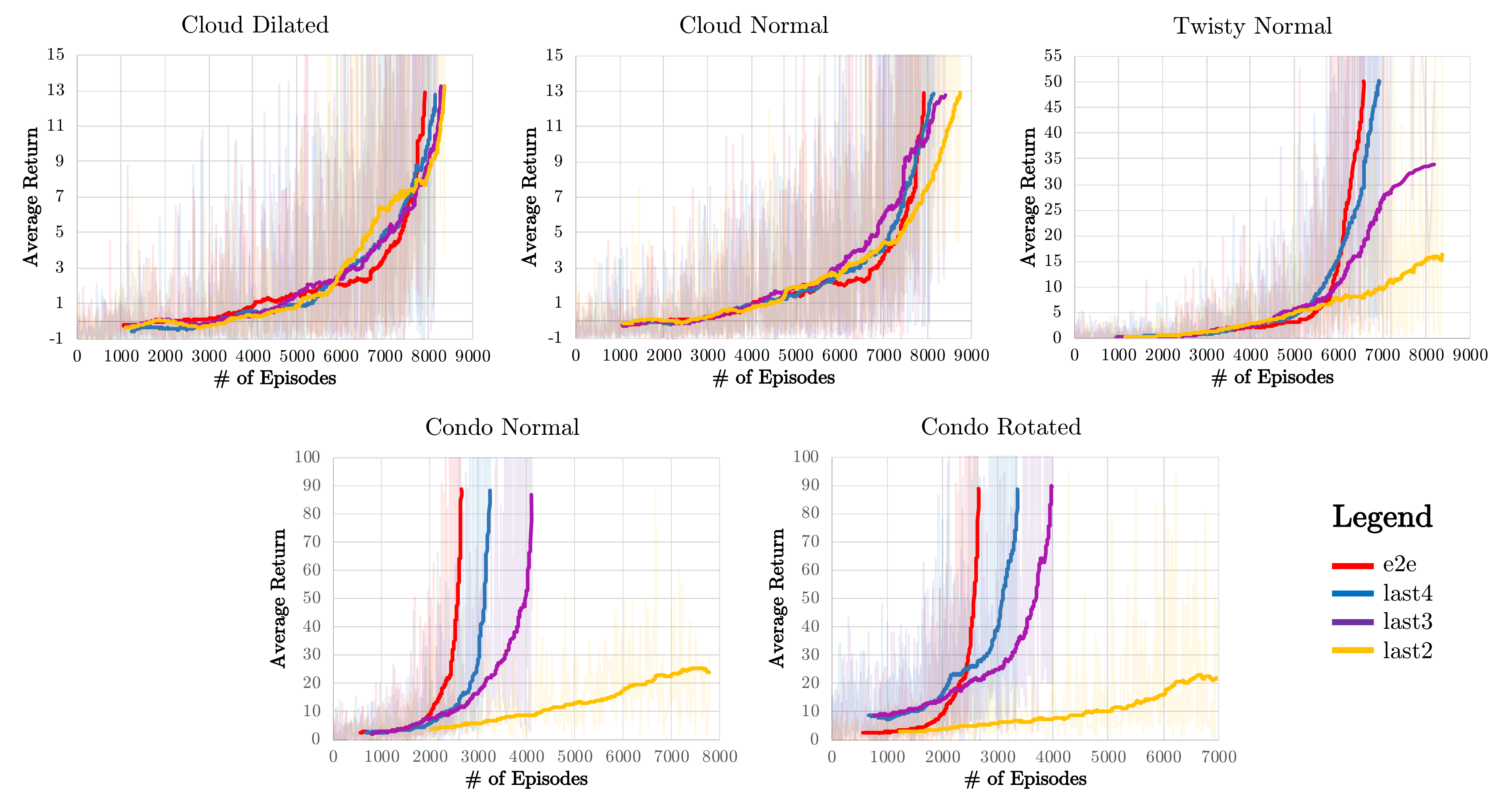}
\caption{Return across environment and action space combination}
\label{fig:ret_graph}
\end{figure*}

\section{Experimentation}
The idea is to show that once the network was trained on meta environments, this knowledge can be used to help train another network for a similar but different problem. The similarity of the problem is kept by having the same object, i.e. autonomous navigation, while the `different' part is achieved by changing/varying the environment and action space. This is done to show that this learning approach is robust to variation in environment and agent's control dynamics. The complete training block diagram can be seen in Fig. \ref{fig:drl_train}.

\subsection{Environmental Variation}
Environmental variation was carried out by designing 3 test environments (named Cloud, Condo and Twisty) with variation in the floor plan, lighting and textures as that of used in the meta environments. The floor plans and snapshots at different locations of these test environments can be seen in the Fig \ref{fig:test_envs}. These environments were designed with a varying degree of similarity to the environments in the used for meta-training and will be discussed in next section.

\subsection{Action Space Variation}
Action space variation was carried out by defining 2 other action spaces along with the one used during the meta training phase. The actual action space was dilated and rotated to generate two other action spaces. The explanation of these action spaces can be seen in Fig \ref{fig:action_var}. The dilated action space was created by dilating the yaw and pitch angles in the original action space by 20\%, while the rotated action space rotates the original action space by 25\% for both pitch and yaw. Both of these action spaces were made probabilistic by introducing noise in the angles (pitch and yaw) as explained in the previous section.

\section{Experimental Results}
\begin{figure}[H]
\centering
\includegraphics[width=0.4\textwidth]{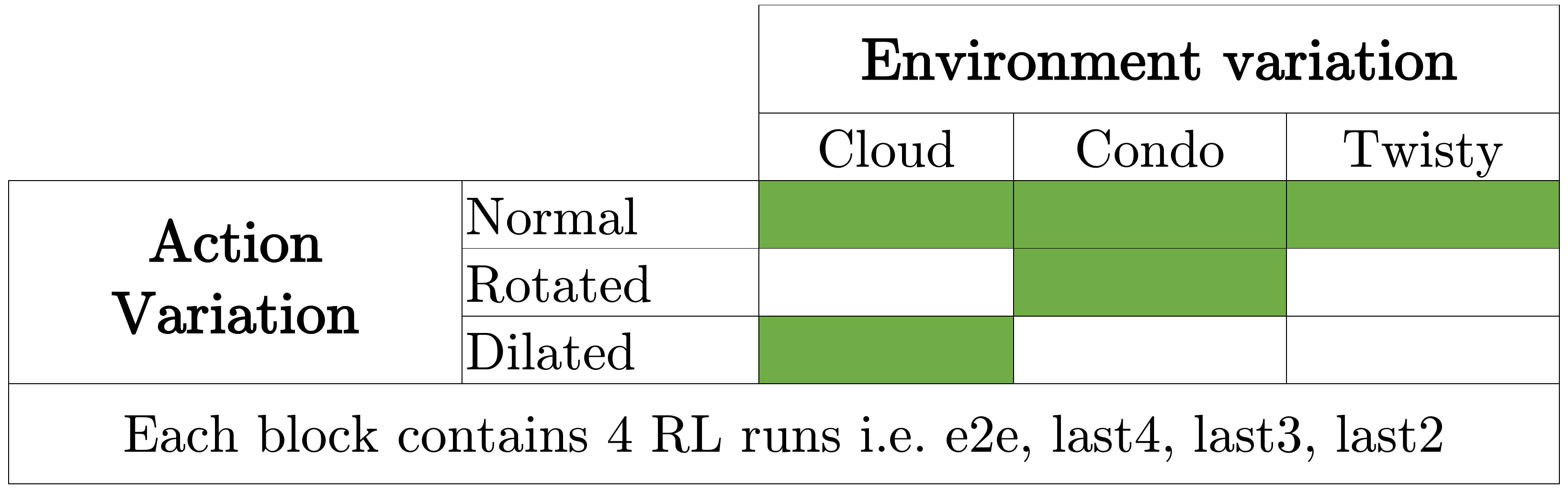}
\caption{Training combination used across the environment and action space variation}
\label{fig:combination}
\end{figure}

In this section we evaluate the proposed approach and quantify the algorithmic performance and computational cost for each train type across different test environments. Experimentation was carried out on a workstation with GTX1080 GPU. As mentioned in the previous section, a list of 20 experimentation was carried out by varying the environment and action space. The list of combination used during experimentation is shown in Fig \ref{fig:combination}

For each of these combinations, the agent was initialized at three different initial position randomly chosen prior to learning. A dueling network was learned using DDQN and PER. The network was first trained end-to-end updating all the weights of the network for 150,000 steps and the return was recorded. The algorithm used for offline training phase can be seen in Algo \ref{alg:E2ERL}. This return serves as a baseline setting a threshold for subsequent train types (last4, last3 and last2). For these train types, the network was trained for either at most 300,000 steps or until the return matched that of e2e train type. 

\begin{figure*}[t]
\centering
\includegraphics[width=0.98\textwidth]{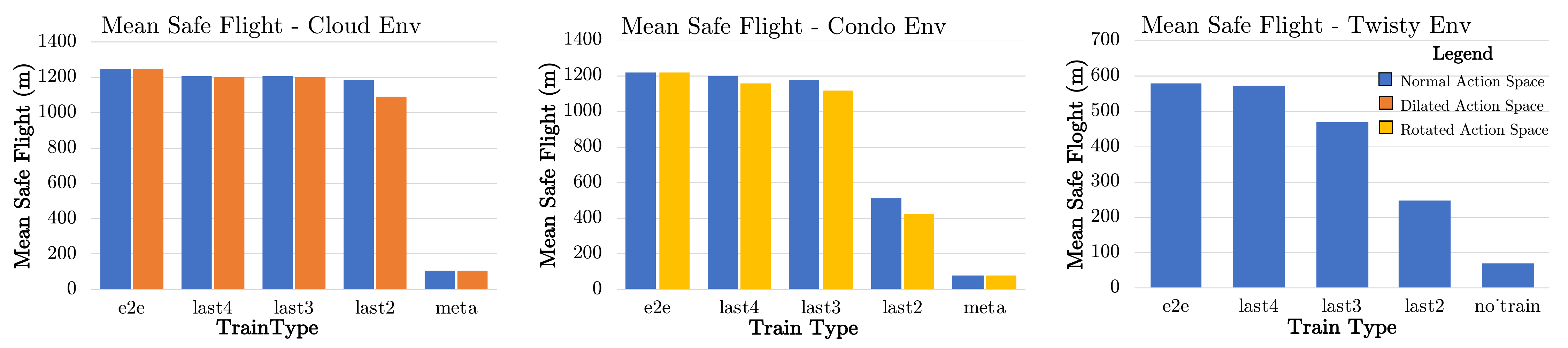}
\caption{Mean Safe Flight (MSF) across different environment for different action spaces.}
\label{fig:msf}
\end{figure*}

\subsection{Algorithmic Performance}
The return graph for all these combinations has been plotted in Fig \ref{fig:ret_graph}. The return graph reported/plotted is the moving average of the actual return graph to make it more meaningful. It can be seen that in all the cases, \textit{train type} last4, and for some cases others, were able to match the return obtained for the train type e2e. It should be noted that variations in action space didn't bar the network to achieve the required return. The only difference that it made was the time/steps required to achieve that return. It took slightly longer to achieve the desired return.

\subsubsection{Test environment 1 - Cloud}
This environment had a smooth floor plan (no sharp edges) and all the wall textures used in this environment were chosen from the 40 texture pool used in the construction of meta environment. Amount of learning transferred from meta environments to this test environment should be significant due its greater similarity to meta environments. This can be seen in the return graph for this environment as shown in Fig \ref{fig:ret_graph}. Not only did all the train types were able to reach the desired return value, but they also did it in almost equal number/amount of iterations/time. 

\subsubsection{Test Environment 2 - Condo}
The floor plan of this environment had turns similar to that of meta environment. 75\% of the textures used for the walls were chosen from the 40 textures pool used during the design of the meta environments. Rest of the 25\% textures were the ones that were never used in the meta environments. The idea is to evaluate the robustness of the approach to variation in the environment characteristics. The idea is to evaluate the performance of the approach to unseen textured environment. The return graph for different train types can be seen in Fig \ref{fig:ret_graph}. It can be seen that except for the train type last2, all the other train types were able to achieve the desired return value. Since this environment has lesser similarity with meta environment as compared to that of Cloud environment, the train type last4 and last3 took longer to achieve the desired return value.

\subsubsection{Test Environment 3 -  Twisty}
Half or the textured used in this environment was new and had never used in the design of meta environments. The floor plan has sharp turns and narrower hallways as compared to other environments. Only train type last4 was able to achieve the desired return threshold, while last 3 performed better than last 2. The respective return graph can be seen in Fig \ref{fig:ret_graph}.

\begin{table}
\centering
\caption{Mean Safe Flight (MSF)}
\begin{tabular}{|c|c|c|c|c|c|c|} 
\hline
\multicolumn{7}{|c|}{\textbf{Mean Safe Flight (m)}}                                         \\ 
\hline
\textbf{Env }  & \textbf{$A_s$} & \textbf{e2e}& \textbf{last4}  & \textbf{last3}  & \textbf{last2}  & \textbf{meta}   \\ 
\hline
\multirow{2}{*}{Cloud} & Normal       & 1245.7 & 1209.0 & 1206.5 & 1187.5 & 110.0  \\
                       & Dilated      & 1245.7 & 1203.0 & 1197.0 & 1093.0 & 110.0  \\ 
\hline
\multirow{2}{*}{Condo} & Normal       & 1218.5 & 1196.6 & 1175.1 & 512.1  & 77.8   \\
                       & Rotated      & 1218.5 & 1153.4 & 1118.0 & 425.0  & 77.8  \\ 
\hline
Twisty                 & Normal       & 580.7  & 573.2  & 468.4  & 248.0  & 68.8   \\
\hline
\end{tabular}
\label{tab:msf}
\end{table}

Mean Safe Flight (MSF) was used to meaning-fully quantify the performance of the learned networks in the respective environment. MSF is the average distance traveled by the agent, in meters, before a collision. For each of the learning combination, the network was initialized with the learned weights and the agent was initialized randomly at 10 different locations within the environment. In order to have a fair comparison, the agent was placed exactly the same way (in terms of position and orientation) across all the train types. In each of the cases, the distance traveled by the agent before collision was recorded and averaged out to generate the MSF. These actual MSF values can be seen in the table \ref{tab:msf} and the normalized MSF value for each environment is plotted in figure \ref{fig:msf}. Th right most column `meta' shows the MSF values achieved by the network initialized with meta-weights without fine-tuning.
It can be seen that for all the cases, the MSF achieved by the train type last4 is at least 97\% that of achieved by end-to-end training. MSF achieved by all the train types co-relates with their return values. 

\begin{table*}[]
\centering
\caption{GPU parameters for different train types}
\begin{tabular}{|l|c|c|c|c|c|}
\hline
\textbf{Train Type} & \multicolumn{1}{l|}{\textbf{Runtime(s)}} & \multicolumn{1}{l|}{\textbf{DtoD memCpy (MB)}} & \multicolumn{1}{l|}{\textbf{GPU Mem (MB)}} & \multicolumn{1}{l|}{\textbf{GPU Load (\%)}} & \multicolumn{1}{l|}{\textbf{Energy/iter (J)}} \\ \hline
e2e                 & 40.59617                                 & 586.3                                          & 4389.0                                     & 0.40188                                     & 5.87                                          \\
last4               & 24.01384                                 & 254.3                                          & 3364.0                                     & 0.21432                                     & 1.85                                          \\
last3               & 23.17413                                 & 220.7                                          & 3362.0                                     & 0.20457                                     & 1.71                                          \\
last2               & 22.67000                                 & 203.9                                          & 3298.0                                     & 0.19234                                     & 1.57                                          \\ \hline
\end{tabular}
\label{tab:gpu_param}
\end{table*}

\begin{figure}[t]
\centering
\includegraphics[width=0.44\textwidth]{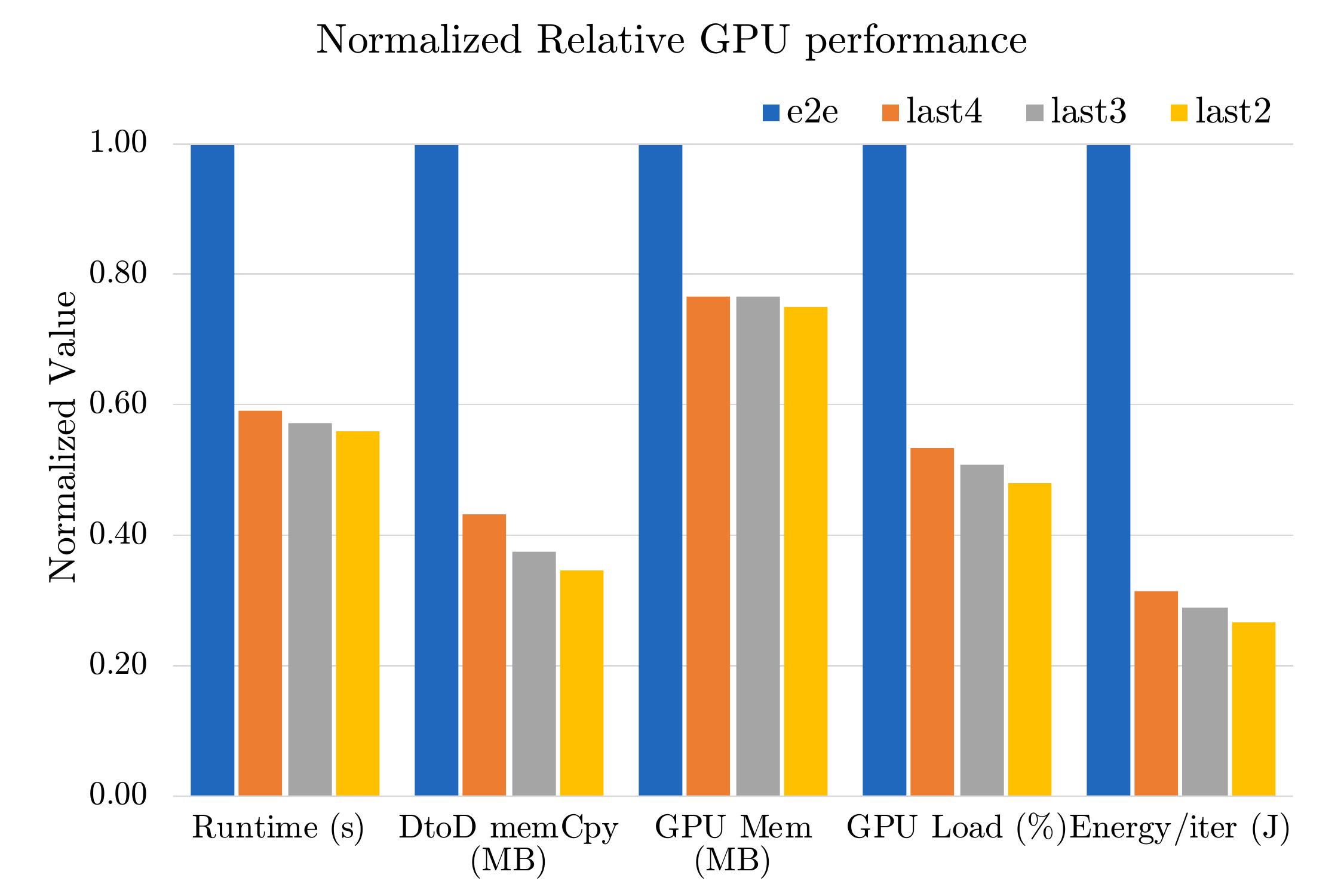}
\caption{GPU parameters for the 4 different train types}
\label{fig:gpu_param}
\end{figure}

\subsection{Computational Cost}\label{sec:compute_cost}
To measure the resources used during training, for each of the train type,  a set of GPU parameters were recorded. These computational parameters were collected using NVIDIA’s profiling tools (nvidia-smi \cite{nvidia_smi} and nvprof \cite{nvprof}) and include
\begin{itemize}
    \item \textbf{Runtime}: 	Time in seconds taken to train the neural network for $K$ iterations
    \item \textbf{DtoDMemcpy}: 	Amount of data transferred (in MBs) within the GPU cores
    \item \textbf{GPU Mem}: 	Amount of GPU Memory used
    \item\textbf{GPU Load}: 	Power consumption of GPU in Watts
    \item \textbf{Energy/iter}: 	Energy consumption per training iteration  
\end{itemize}
Runtime and GPU load corresponds to latency and power required for training, while DtoDMemcpy and GPU Mem governs the hardware resources required. These parameters give a quantitative way of understanding how these different train types directly affects the edge node resources. In order to calculate these parameters, for each train type, the neural network was trained for $K=500$ number of iterations on a collected data-set. These GPU parameters have been tabulated in Table \ref{tab:gpu_param} and their normalized values have been plotted in Fig. \ref{fig:gpu_param}. It can bee seen that for all the train types the time required to train the network (latency) was reduced to less than 60\% as compared to that of e2e while reducing the energy consumption to less than 30\%. The reduced latency directly dictates the speed of the drone during training. 

For a given speed of the drone, the corresponding distance traveled between two sequentially acquired frames, and the drone distance threshold for obstacles (a measure of clutter in the environment), we can calculate the minimum number of Frames per Second (FPS) required for collision avoidance. For a drone to have a higher speed, it needs to be able to process more frames in a given amount of time (i.e. support higher FPS). The drone will only be able to support that speed if the underlying computational system can process the dictated FPS (which is inverse of the per frame latency). So, the maximum speed of the drone will be limited by the latency of the system. Hence the latency improvement of last2 vs e2e in Fig \ref{fig:gpu_param} directly corresponds to an improvement of maximum supported theoretical speed (based purely on the training pass and ignoring other latency sources) of about 1.8 times from e2e to last2.
Using the lower train types not only reduces the latency but also requires less operating power. Since it was reported in Fig \ref{fig:msf} that the algorithmic performance (in terms of MSF) for these different train types was comparable to e2e learning, reduced hardware, power and time requirement makes it favorable to be implemented on edge nodes.

\begin{figure}[t]
\centering
\includegraphics[width=0.48\textwidth]{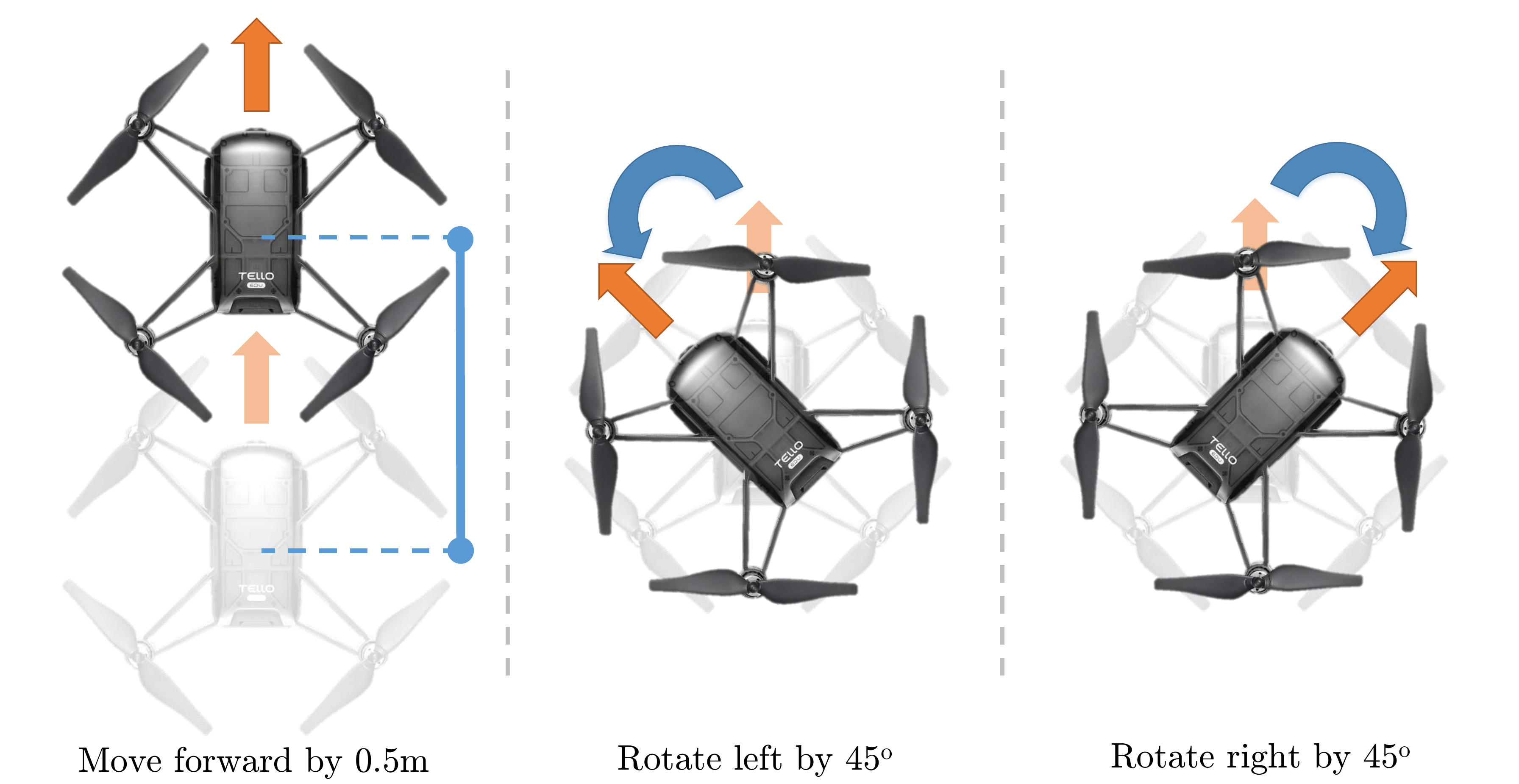}
\caption{Action space of Tello drone for real environment}
\label{fig:tello_actions}
\end{figure}

\begin{figure}[t]
\centering
\includegraphics[width=0.48\textwidth]{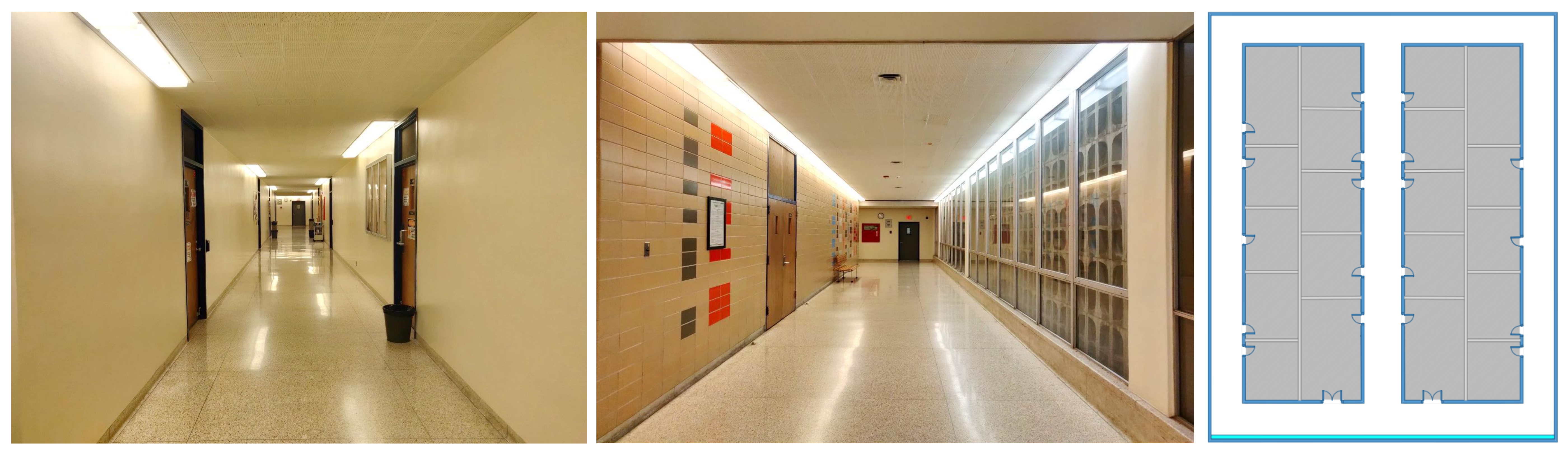}
\caption{Snapshots and the layout of the Hallway arena used as test real environment}
\label{fig:hallway}
\end{figure}

\begin{figure*}[t]
\centering
\includegraphics[width=0.98\textwidth]{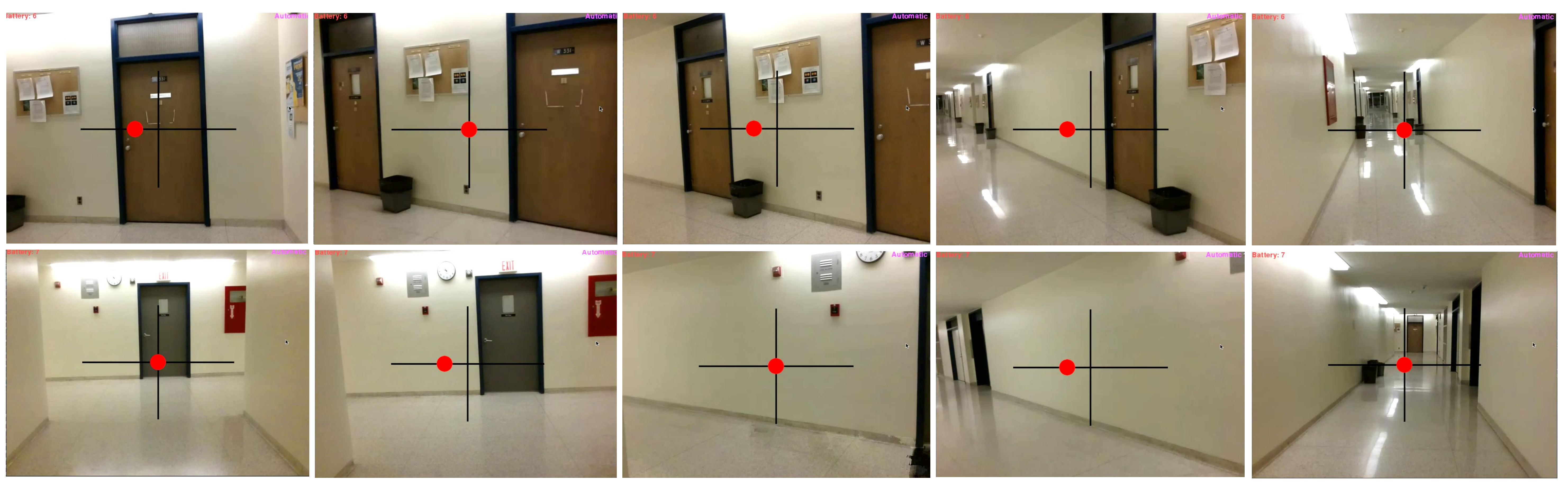}
\caption{Action predictions by the network for the Hallway environment}
\label{fig:hallway_pred}
\end{figure*}

\begin{table*}[t]

\centering
\begin{tabular}{|c|c|c|c|c|}
\hline
\textbf{Arena}                       & \textbf{Train Type} & \textbf{MSF (m)} & \textbf{Normalized Energy} & Normalized GPU Load\\ \hline
\multirow{2}{*}{Hallway} & E2E (\textsc{NavREn}-RL)       & 16.1      & x1 &           x1                 \\
                                     & Last2 (This work) & 15.6         & x0.27 &        x0.48           \\ \hline

\end{tabular}
\caption{MSF on real hallway environment}
\label{table:stat_real}
\end{table*}

\subsection{Experimental Verification with DJI Drone in Real Environment}\label{real_drone}
Experimentation with real drone in a real environment was carried out by transferring the learning from a simulated meta-environment. A low cost DJI Tello drone was used for this real-time experimentation. DJI Tello does not have the computational power to carry out the required processing on-board. Hence, a workstation/cloud equipped with a core i7 processor and GTX1080 GPU was used for training. Tensorflow was used as the ML platform to carry out the neural network computation on the workstation. 

Offline training was carried out on the same simulated meta-environments (Fig \ref{fig:test_envs}) for the same modified AlexNet network (Fig. \ref{fig:cnn}). The action space, however was modified to contain only three actions. These actions include going forward by 0.5m, rotating clock-wise by 45 degrees, and rotating counter clock-wise by 45 degrees and can be seen in Fig. \ref{fig:tello_actions}. The action space did not include any actions that corresponds to changing the drone altitude. Once the network was trained for the three-action action space on the simulated meta-environments, the learned weights were used as initializers for the network to be trained in a real environment. For this purpose a hallway environment of an engineering building was used that contains glass walls and corridors $\sim 1.5m$ wide and can be seen in Fig \ref{fig:hallway}.

Using the baseline deep reinforcement learning algorithm in a real environment is time-consuming. Hence the approach discussed in \cite{anwar2018navren} was used. Using this approach, an expert user collects a set of data-points in the real environment. These expert data-points are made a mandatory part of the experience replay from which the data-points are sampled for training. Moreover data-aggregation techniques are used when the drone virtually crashes to aid the data-collection. Only the last two layers of the network were updated during training, while the weights in the rest of the layers were kept static.

Once the network was trained for the last 2 layers, the drone was placed at different initial positions and the performance of the network was observed. MSF was used as the performance metric. Fig \ref{fig:hallway_pred} shows the control actions predicted by the network for the given camera frames. Table \ref{table:stat_real} reports the Mean Safe Flight (MSF) of the drone in the Hallway environment when using the network that was trained for its last 2 layers and compares it with the MSF of the network which was trained end-to-end. It can be seen that both the networks give similar performance in terms of MSF while the GPU load was reduced by 2 times.

\section{Conclusions}
This paper implements a Transfer learning approach to reduce the amount of resources required to train a deep neural network for RL problem by training the network on a set of rich and diverse meta environments, transferring the domain knowledge to test environments and training the last few fully connected layers only. The algorithmic performance of this network measured in terms of Mean Safe Flight was similar to training the network end-to-end while reducing the latency and energy consumption by 1.8 and 3.7 times respectively. The reduction in these parameters can make it possible for DRL training to implemented resource constrained edge nodes. Moreover, the approach was tested on a real environment using a low cost drone and showed similar performance.



\section*{Acknowledgments}

This work was supported in part by C-BRIC, one of six centers in JUMP, a Semiconductor Research Corporation (SRC) program sponsored by DARPA.

    
\bibliographystyle{ieeetran}
\bibliography{references}

\begin{figure*}[p]
\centering
\includegraphics[width=0.98\textwidth]{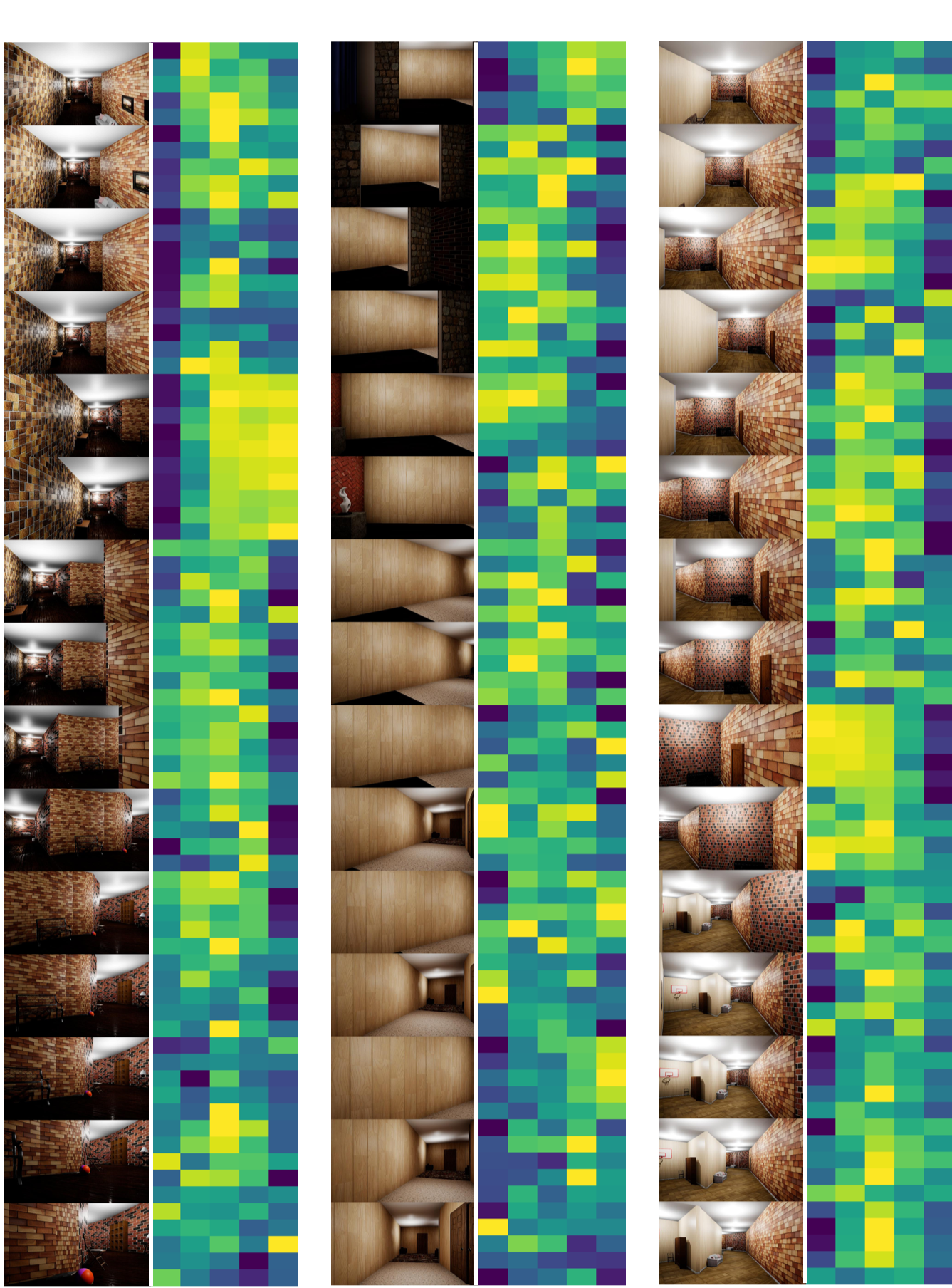}
\caption{Images captured from front facing camera of the drone during flight in simulated environments. On the right of each block is the action space probability where blue corresponds to lower and yellow higher probability.  From left to right: Cloud, Condo and Twisty environment}
\label{fig:flight_graph}
\end{figure*}

\appendix
Fig \ref{fig:flight_graph} shows the images captured from the front facing camera of the drone during flight across the three different \textbf{simulated} test environments. For each environment, the RGB image of the camera (on the left) and the $5\times5$ network predicted action space (on the right) has been shown. Each of the bin in the predicted action space represents the normalized Q values (across all the predictions). The darker (blue) bins corresponds to smaller values while lighter (yellow) corresponds to higher Q values. Moving in the direction of darker bins will increase the probability of collision.

\end{document}